\documentclass[10pt,twocolumn,letterpaper]{article}

\usepackage{cvpr}              

\usepackage{graphicx}
\usepackage{amsmath}
\usepackage{amssymb}
\usepackage{booktabs}
\usepackage{array}
\usepackage{multirow}
\usepackage{pifont}
\usepackage{color}
\newcommand{\cmark}{{\ding{51}}}
\newcommand{\xmark}{{\ding{55}}}
\newcommand{\authorskip}{\hspace{2.5mm}}
\usepackage{adjustbox}
\usepackage{threeparttable}
\usepackage[pagebackref,breaklinks,colorlinks]{hyperref}

\usepackage[capitalize]{cleveref}
\crefname{section}{Sec.}{Secs.}
\Crefname{section}{Section}{Sections}
\Crefname{table}{Table}{Tables}
\crefname{table}{Tab.}{Tabs.}

\def\cvprPaperID{10608} 
\def\confName{CVPR}
\def\confYear{2022}

\begin{document}

\title{Portrait Interpretation and a Benchmark}

\author{
 Yixuan Fan \authorskip Zhaopeng Dou \authorskip Yali Li \authorskip Shengjin Wang \\[2mm]
 Tsinghua University\\[2mm]
 {\tt\small \{fan-yx21, dcp19\}@mails.tsinghua.edu.cn, \{liyali13, wgsgj\}@tsinghua.edu.cn}
}
\maketitle

\begin{abstract}

We propose a task we name \emph{Portrait Interpretation} and construct a dataset named \emph{Portrait250K} for it. Current researches on portraits such as human attribute recognition and person re-identification have achieved many successes, but generally, they: 1) may lack mining the interrelationship between various tasks and the possible benefits it may bring; 2) design deep models specifically for each task, which is inefficient; 3) may be unable to cope with the needs of a unified model and comprehensive perception in actual scenes. In this paper, the proposed portrait interpretation recognizes the perception of humans from a new systematic perspective. We divide the perception of portraits into three aspects, namely Appearance, Posture, and Emotion, and design corresponding sub-tasks for each aspect. Based on the framework of multi-task learning, portrait interpretation requires a comprehensive description of static attributes and dynamic states of portraits. To invigorate research on this new task, we construct a new dataset that contains 250,000 images labeled with identity, gender, age, physique, height, expression, and posture of the whole body and arms. Our dataset is collected from 51 movies, hence covering extensive diversity. Furthermore, we focus on representation learning for portrait interpretation and propose a baseline that reflects our systematic perspective. We also propose an appropriate metric for this task. Our experimental results demonstrate that combining the tasks related to portrait interpretation can yield benefits. Code and dataset will be made public.

\end{abstract}
\section{Introduction}
\label{sec:intro}

\begin{figure}[t]
  \centering
  \includegraphics[width=1.0\linewidth]{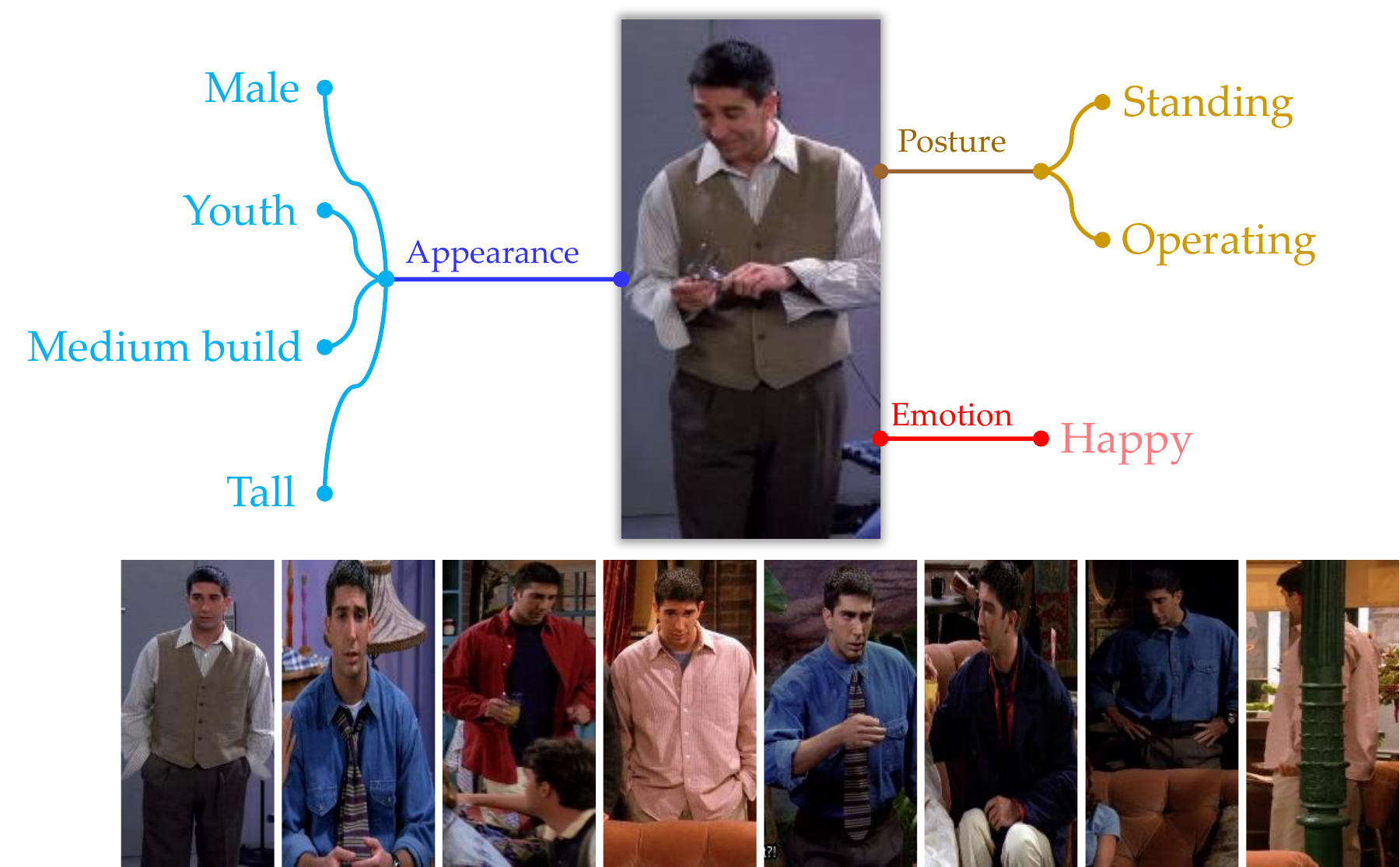}
  \caption{Portrait interpretation generates a comprehensive perception of the target while associating the given image with others of the same person.}
  \label{fig:demo}
\end{figure}

Humans are born with the ability of comprehensive perception, especially towards other people. Imagine that we meet a friend on the street, subconsciously we immediately complete the perception: Are they male or female? How old are they? How are their body shapes? At the same time, we notice their dynamic states, such as movements and postures. Besides, we can tell what they are feeling from their facial gestures. And most importantly, we identify who they are, that is, to connect them with the friend in our memory. 

Since the era of deep learning, the ability of machines to tackle many specific tasks has far surpassed that of humans. For example, facial recognition systems can distinguish hundreds of millions of people. On the other hand, these abilities are not exactly the same as the abilities needed to develop a human-like agent. Towards the goal of artificial general intelligence, many attempts are also underway. In this paper, we focus on the ability of comprehensive perception and propose the task of \textbf{Portrait Interpretation}, trying to transfer the abilities that humans show when they meet friends on the street to the machine. In the field of image segmentation, tasks based on the idea of comprehensive perception include panoptic segmentation \cite{kirillov2019panoptic}, which combines semantic and instance segmentation. In the intersection of computer vision and natural language processing, we believe that visual question answering \cite{antol2015vqa} has similar characteristics. 

Starting from the vivid example of meeting a friend, we find that existing tasks related to this process include pedestrian attribute recognition and person re-identification (re-id). The question then arises: \emph{In what way should we combine these tasks?} By introducing portrait interpretation, we provide a systematic perspective. 

Portrait interpretation divides the perception towards humans into three aspects, namely Appearance, Posture, and Emotion, and incorporates several sub-tasks for each aspect, including various attribute recognition tasks. Reasonably, re-id belongs to the perception of appearance. Figure \ref{fig:demo} shows our design, recognizing a variety of attributes and re-id are completed at the same time. 

To promote research in portrait interpretation, we construct a large-scale dataset named \textbf{Portrait250K} that can serve as a benchmark. We collect 250,000 portraits from 51 movies and TV series across countries, and manually label them with eight kinds of labels, corresponding to eight sub-tasks. The images and the distribution of labels show many characteristics that naturally exist in the real world, including but not limited to long-tailed or unbalanced distribution, variable occlusion, truncation, illumination, and changes in clothing, makeup, as well as background environment of the characters.

To measure the performance of models, we not only design metrics for each sub-task, but also propose a unified metric named \textbf{Portrait Interpretation Quality} (PIQ), which reflects our systematic perspective, and reasonably assign weights for metrics of each sub-task.

Furthermore, we design a simple baseline for portrait interpretation under the paradigm of multi-task learning. We focus on multi-task representation learning and propose a feature space split scheme. We also propose a simple metric learning loss. Through experiments, we prove the feasibility and superiority of the research of portrait interpretation.
\section{Related Work}
\label{sec:related-work}

\begin{table*}
  \centering
  \small
  \begin{tabular}{l|cc|cccc|cc|c}
        \toprule
        \multirow{2}{*}{Datasets} & \multirow{2}{*}{\# of Images} & \multirow{2}{*}{\# of IDs} & \multicolumn{4}{c|}{\textbf{Appearance}} & \multicolumn{2}{c|}{\textbf{Posture}} & \textbf{Emotion} \\
        ~ & ~ & ~ & Gender & Age & Physique & Height & Body & Arm & Expression \\ \midrule
        Berkeley \cite{bourdev2011describing} & 8,035 & \xmark & \cmark & \xmark & \xmark & \xmark & \xmark & \xmark & \xmark \\
        HAT \cite{sharma2011learning} & 9,344 & \xmark & \cmark & \cmark & \xmark & \xmark & \cmark & \xmark & \xmark \\
        APiS \cite{zhu2013pedestrian} & 3,661 & \xmark & \cmark & \xmark & \xmark & \xmark & \xmark & \xmark & \xmark \\
        PETA \cite{deng2014pedestrian} & 19,000 & 8,705 & \cmark & \cmark & \xmark & \xmark & \xmark & \xmark & \xmark \\
        RAPv1 \cite{li2015deepmar} & 41,585 & \xmark & \cmark & \cmark & \cmark & \xmark & \xmark & \cmark & \xmark \\
        CRP \cite{hall2015fine} & 27,454 & \xmark & \cmark & \cmark & \cmark & \xmark & \xmark & \xmark & \xmark \\
        PARSE-27k \cite{sudowe2015person} & 27,000 & \xmark & \cmark & \xmark & \xmark & \xmark & \xmark & \xmark & \xmark \\
        WIDER \cite{li2016human} & 57,524 & \xmark & \cmark & \xmark & \xmark & \xmark & \xmark & \xmark & \xmark \\
        PA-100K \cite{liu2017hydraplus} & 100,000 & \xmark & \cmark & \cmark & \xmark & \xmark & \xmark & \xmark & \xmark \\
        RAPv2 \cite{li2018richly} & 84,928 & 2,589 & \cmark & \cmark & \cmark & \xmark & \xmark & \cmark & \xmark \\
        Market-1501 \cite{lin2019improving} & 33,060 & 1,501 & \cmark & \cmark & \xmark & \xmark & \xmark & \xmark & \xmark \\
        DukeMTMC-reID \cite{lin2019improving} & 36,411 & 1,404 & \cmark & \xmark & \xmark & \xmark & \xmark & \xmark & \xmark\\
        UAV-Human \cite{li2021uav} & 22,263 & 1,144 & \cmark & \xmark & \xmark & \xmark & \cmark & \cmark & \xmark \\ \midrule
        \textbf{Portrait250K} & 250,000 & 1,066 & \cmark & \cmark & \cmark & \cmark & \cmark & \cmark & \cmark \\ \bottomrule
  \end{tabular}
  \caption{Comparison among our Portrait250K and some datasets for Pedestrian Attribute Recognition.}
  \label{tab:dataset-compare}
\end{table*}

Human-centered research is hot in the field of computer vision and has achieved many important successes in the past few years, together with obtaining many applications in the fields of human-computer interaction and intelligent security, etc. Introducing the new task of portrait interpretation is based on the recognition of these successes, and aims at promoting the community to challenge this more difficult task which combines the independent researches into a whole. Next, we review the related tasks as well as researches regarding multi-task learning, which is the basic framework for solving the task of portrait interpretation. 

\noindent\textbf{Tasks related to portrait.}
A task that is close to portrait interpretation is \emph{Pedestrian Attribute Recognition} (PAR), which aims at predicting attributes of a target person. Existing methods for PAR includes visual attention \cite{liu2017hydraplus,tan2019attention,tang2019improving,wu2020distraction,jia2021spatial}, body parts division \cite{bourdev2011describing} and attributes relationship mining \cite{wang2016cnn,zhao2019recurrent} etc. Although this task has been quite sufficiently studied and discussed, it is still difficult in the cases of changing viewpoints, variable illumination, low resolution, occlusion, and blur, etc. 

The core difference between portrait interpretation and PAR is that we dissect portraits at a higher level, that is, analyze portraits from three relatively independent aspects. In the past ten years, many datasets have been proposed for PAR, but none of them can meet the needs of portrait interpretation. Table \ref{tab:dataset-compare} compares these datasets with our Portrait250K. Our dataset outperforms them w.r.t. the data volume and labeled tasks.

In addition to tasks related to attribute recognition, \emph{Person Re-identification} (re-id) is also one of the sub-tasks of portrait interpretation, which aims at associating images of the same person. In the deep learning era, the typical baseline method for person re-id performs image classification during training to learn distinguishable representation. During inference, the classification layer is discarded, and retrieval is done by measuring the distance between the feature vectors (maybe more than one in complex models) of the query image and the gallery images. Like many other tasks in vision, the key to a strong re-id model is learning good representations, which is confirmed by \cite{luo2019bag,luo2019strong}. Not surprisingly, if the model can get more supervision that can help it learn better representations, such as various attributes of portraits, it will also help improve its performance \cite{lin2019improving}.

\noindent\textbf{Multi-Task Learning.}
Tasks in the real world are interrelated in many cases, and multi-task learning (MTL) attempts to solve multiple tasks simultaneously to obtain better generalization performance. In addition, irrelevant information between tasks also helps to reduce over-fitting. Portrait interpretation pays attention to these relationships between different aspects and sub-tasks. Researches on MTL mainly focus on two sub-problems, namely structural design and optimization method. The essence of designing a network for MTL is to make the sub-nets and parameters shared among different tasks, and mainstream strategies can be roughly divided into two categories, namely hard sharing \cite{zhang2014facial,ma2018modeling,liu2019end,zhou2020pattern,vandenhende2020mti,Hu2021UniT,khattar2021cross} and soft sharing \cite{misra2016cross,gao2019nddr,liang2020model}. The hard sharing methods divide the entire model into two parts: the sub-nets closer to the input are shared by all tasks, and then modules are branched out for each task independently. The soft sharing methods usually have independent modules for each task from end to end, while the modules can interact at different stages of the network.

For multiple losses generated by multi-tasks, the most trivial way to get a total loss is to manually assign weights to them. A more flexible and efficient approach is to calculate weights based on the uncertainty of the tasks \cite{kendall2018multi}. 

\section{Portrait Interpretation}
\label{sec:portrait-status-analysis}

The perception of portraits can be divided into three aspects, namely Appearance, Posture, and Emotion. This division takes static attributes and dynamic states into account. For each aspect, we describe it through several sub-tasks. The details of each aspect are introduced below, but please note that the design of these sub-tasks is flexible. In future research, more existing tasks can be combined to expand the task of portrait interpretation. 

\textbf{Appearance} represents static attributes that will not change within a certain period, which are bound to a person's identity. We use four classification tasks regarding gender, age, physique, and height to characterize appearance, and of course, there is the re-id task. Existing datasets may contain attributes regarding hair and clothes, while we believe that they are not strongly bound to people themselves, thus they are not included. 

\textbf{Posture} represents dynamic states, specifically physical movements. We consider the posture of the whole body and arms, respectively. 

\textbf{Emotion} is largely conveyed by facial expressions. Following \cite{ekman1971constants,matsumoto1992more}, we use seven basic expressions. 
\section{The Portrait250K Dataset}
\label{sec:dataset}

\begin{figure*}[t]
  \centering
  \includegraphics[width=1.0\textwidth]{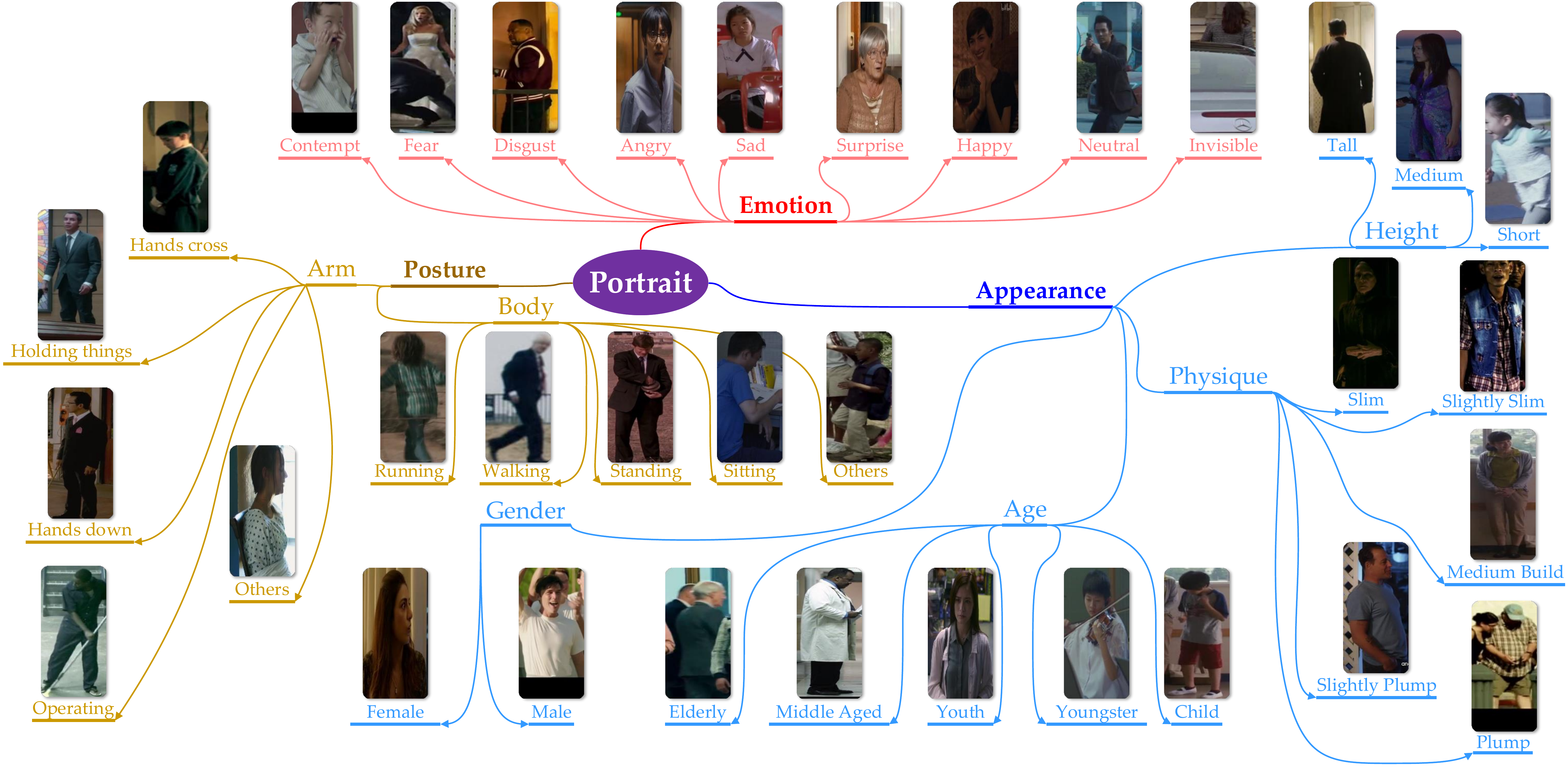}
  \caption{Specific labels used in Portrait250K and their visual examples.}
  \label{fig:label}
\end{figure*}

To invigorate research on portrait interpretation, we construct a large-scale dataset named Portrait250K with rich manual annotations to serve as a benchmark.

\subsection{Data Collection}

We use video data from 51 movies and TV series to construct our dataset. To obtain portrait bounding boxes, we make use of a multi-object tracking model. Multi-object tracking aims at estimating bounding boxes and identities of objects in videos. The JDE model proposed by Wang~\etal~\cite{wang2020towards} jointly outputs the detection results with the corresponding embeddings and resulting in a neat and fast system. For every several frames, we extract bounding boxes output by JDE, and finally 250,000 images with a resolution of $256\times 128$ were obtained.

We recruited a dozen professional annotators to complete the manual annotation of the eight labels regarding these images. In order to reduce the influence of the subjective judgment of different annotators, each annotator only annotates one attribute, thus there will be only one or two annotators for each attribute. Specific labels and corresponding visual examples are shown in Figure \ref{fig:label}.


\begin{figure*}[t]
  \centering
  \includegraphics[width=0.24\textwidth]{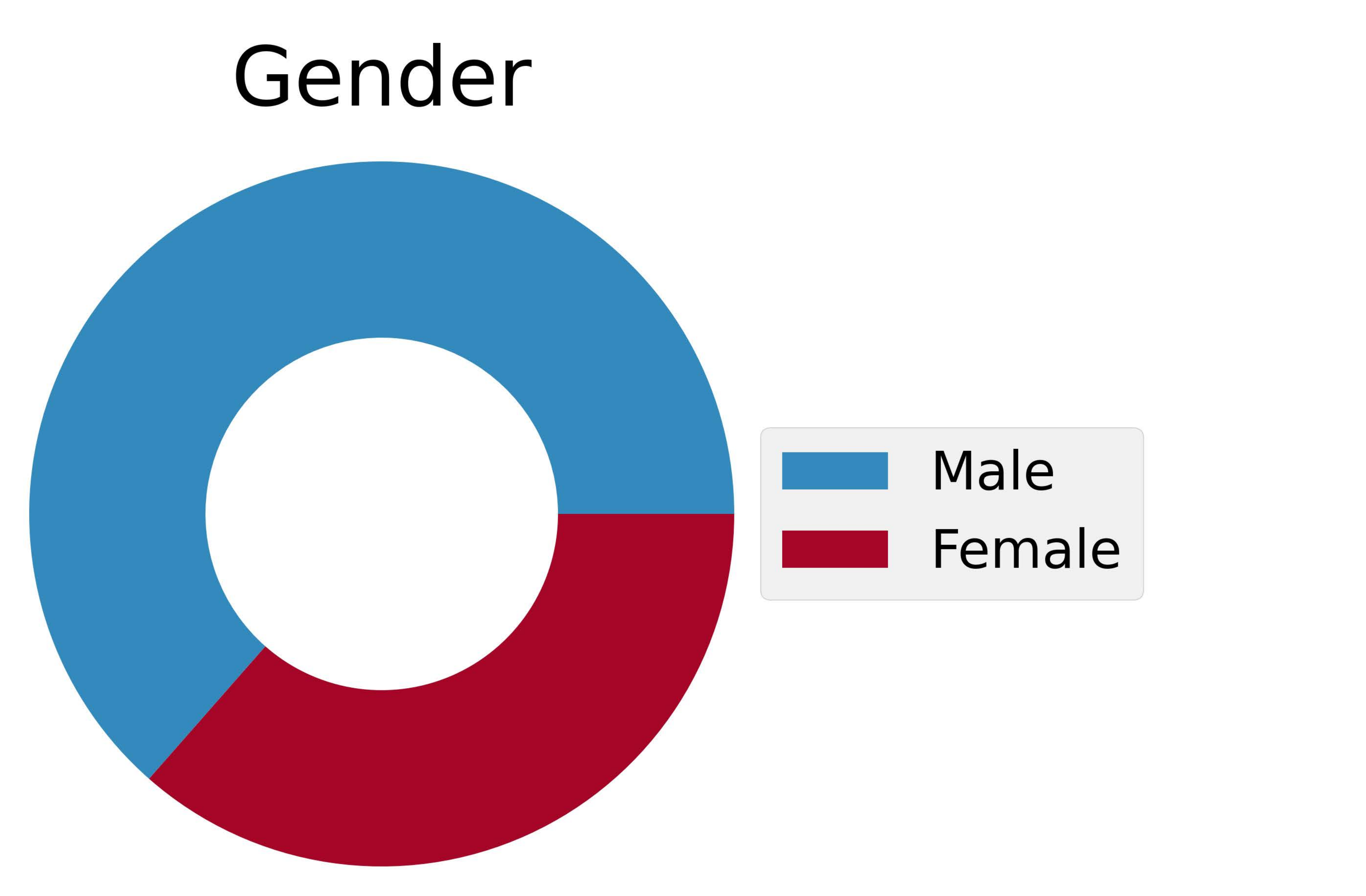}
  \includegraphics[width=0.24\textwidth]{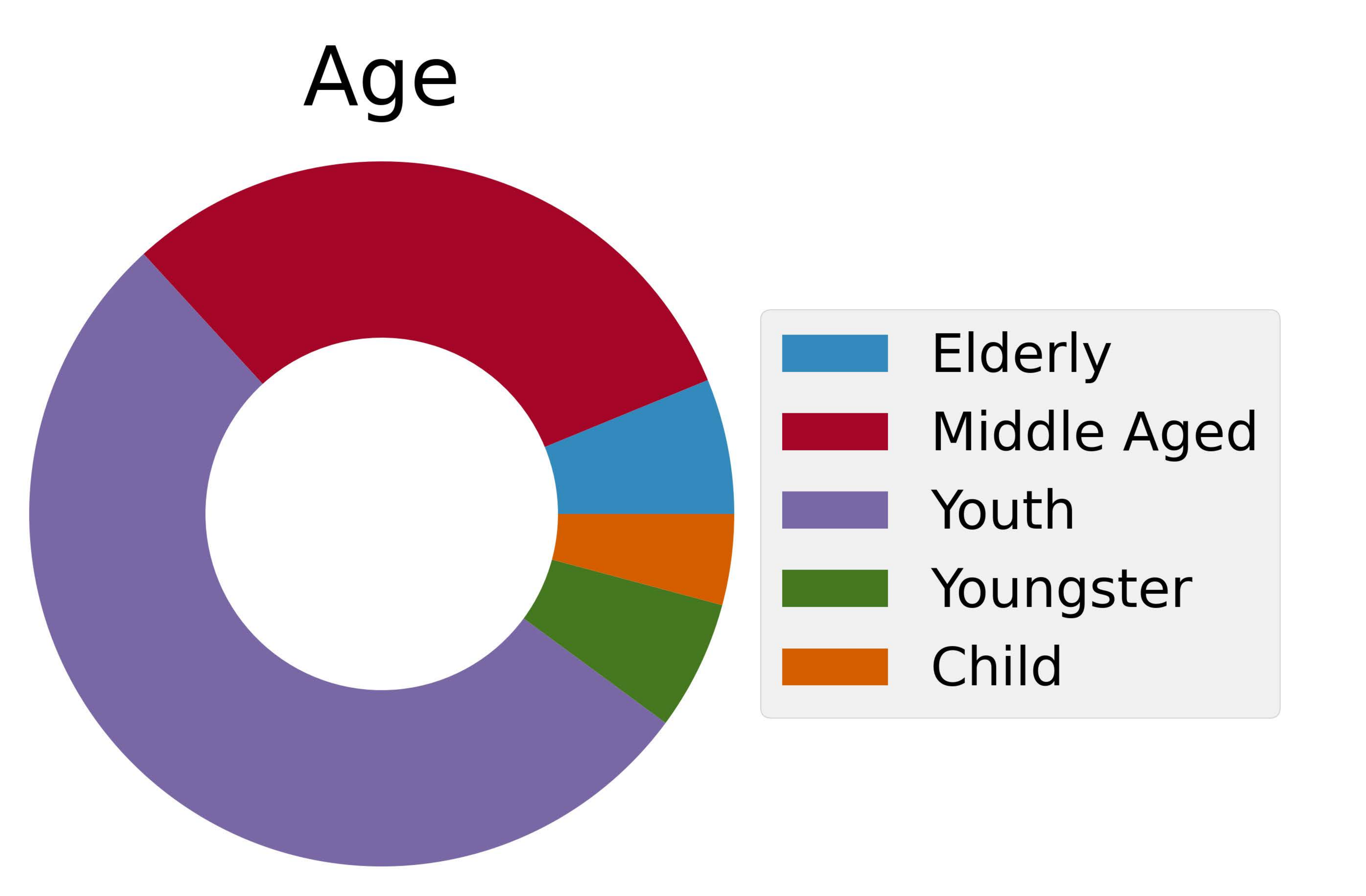}
  \includegraphics[width=0.24\textwidth]{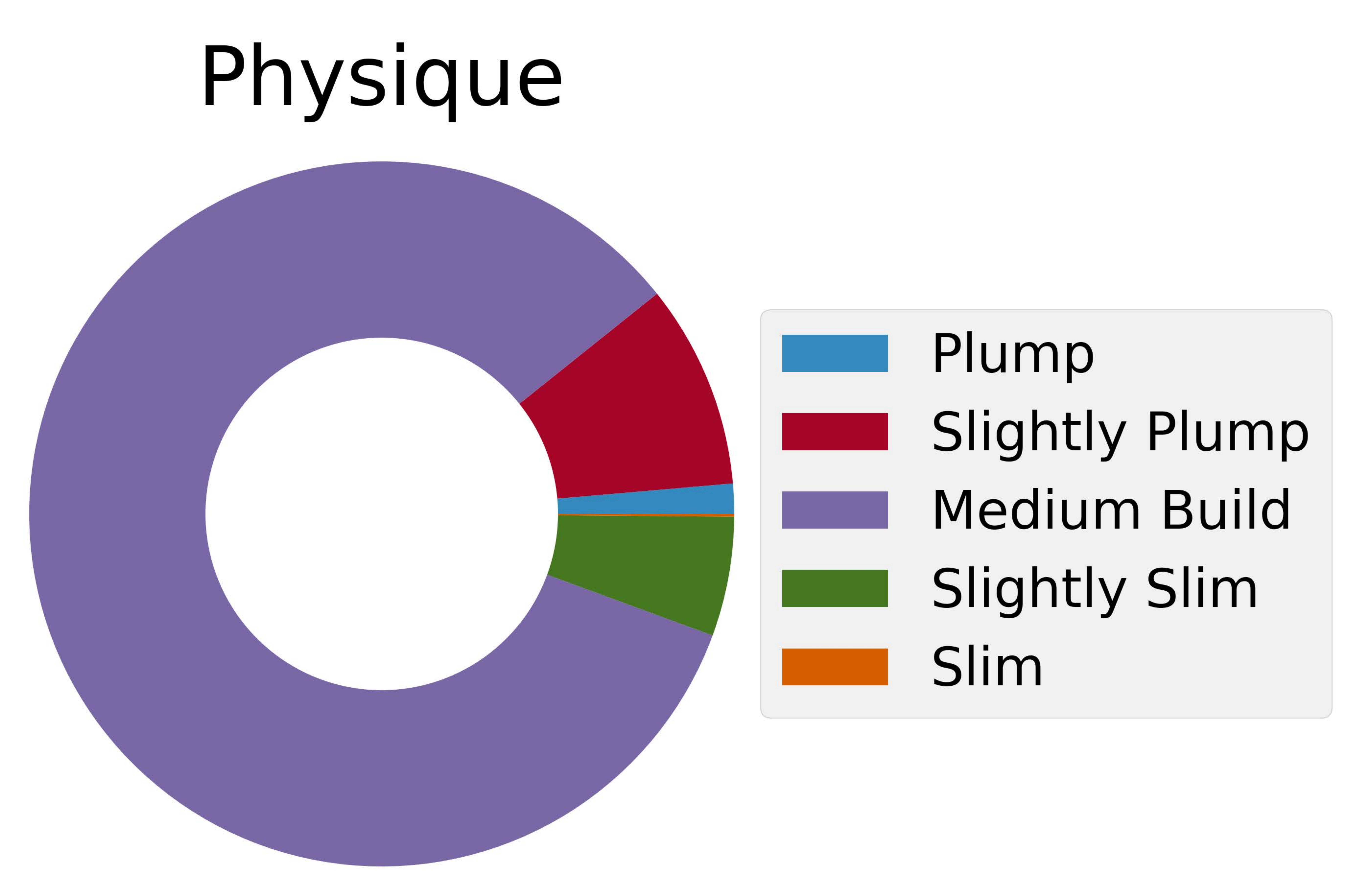}
  \includegraphics[width=0.24\textwidth]{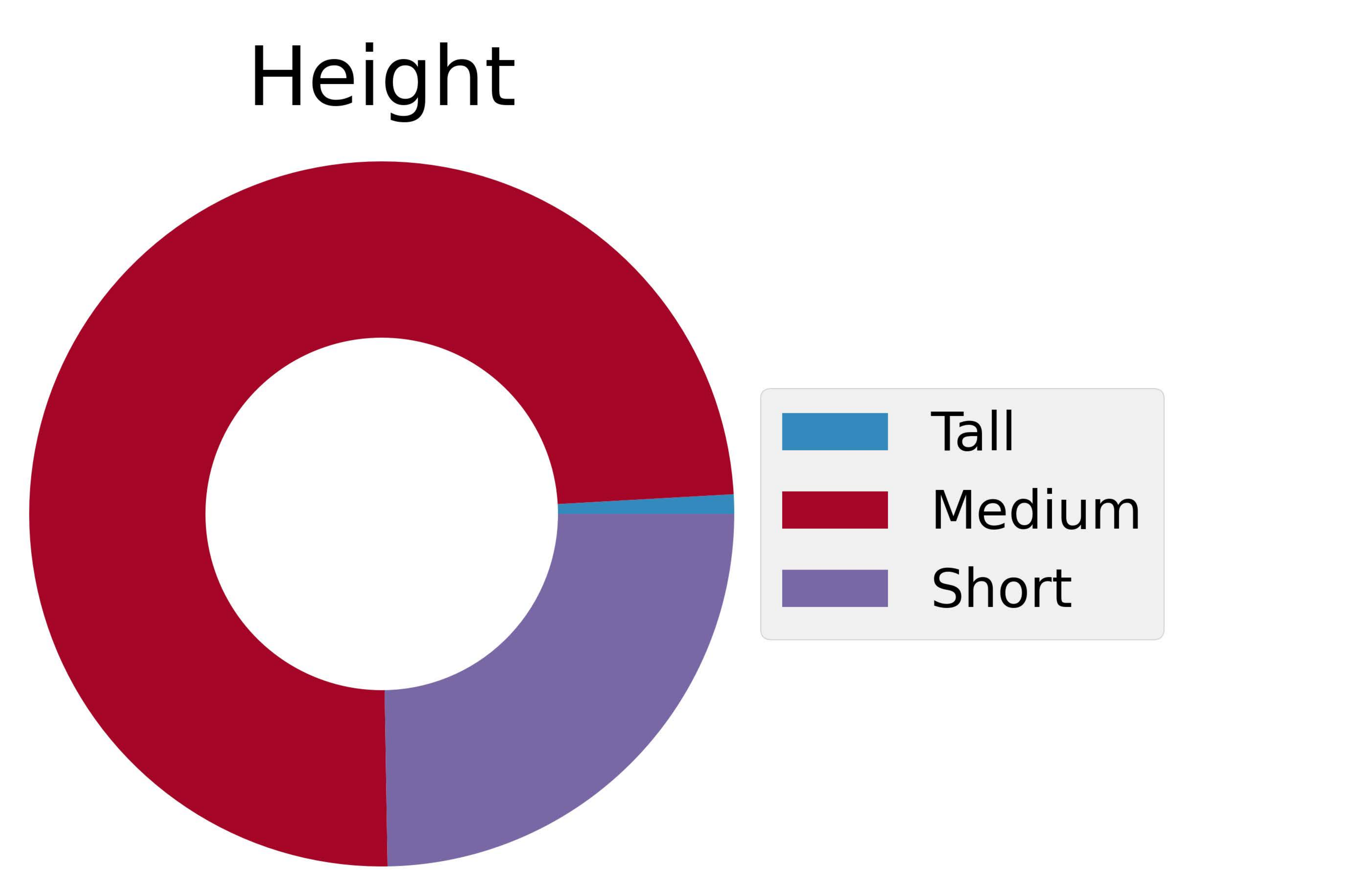}\\
  \includegraphics[width=0.24\textwidth]{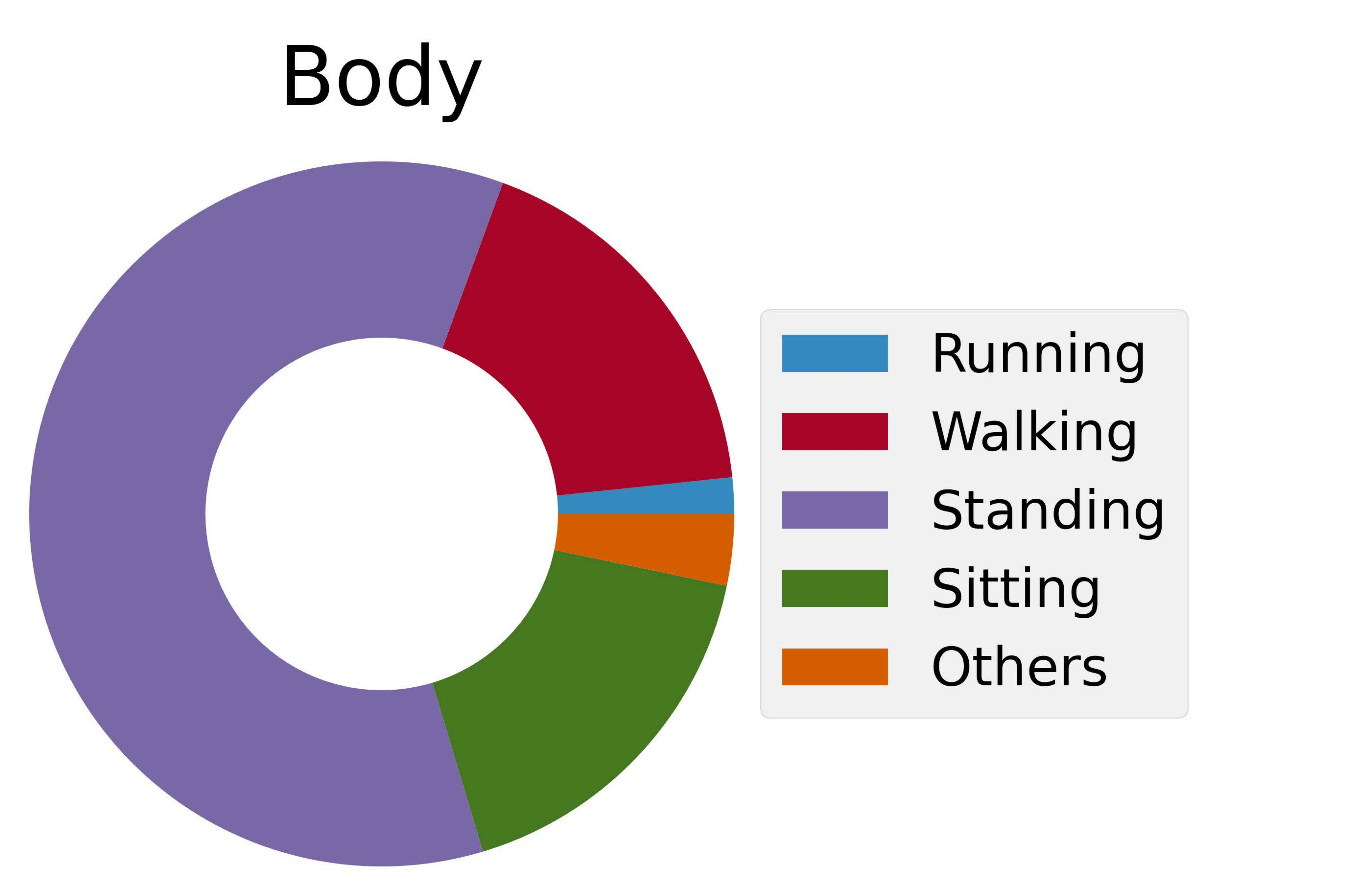}
  \includegraphics[width=0.24\textwidth]{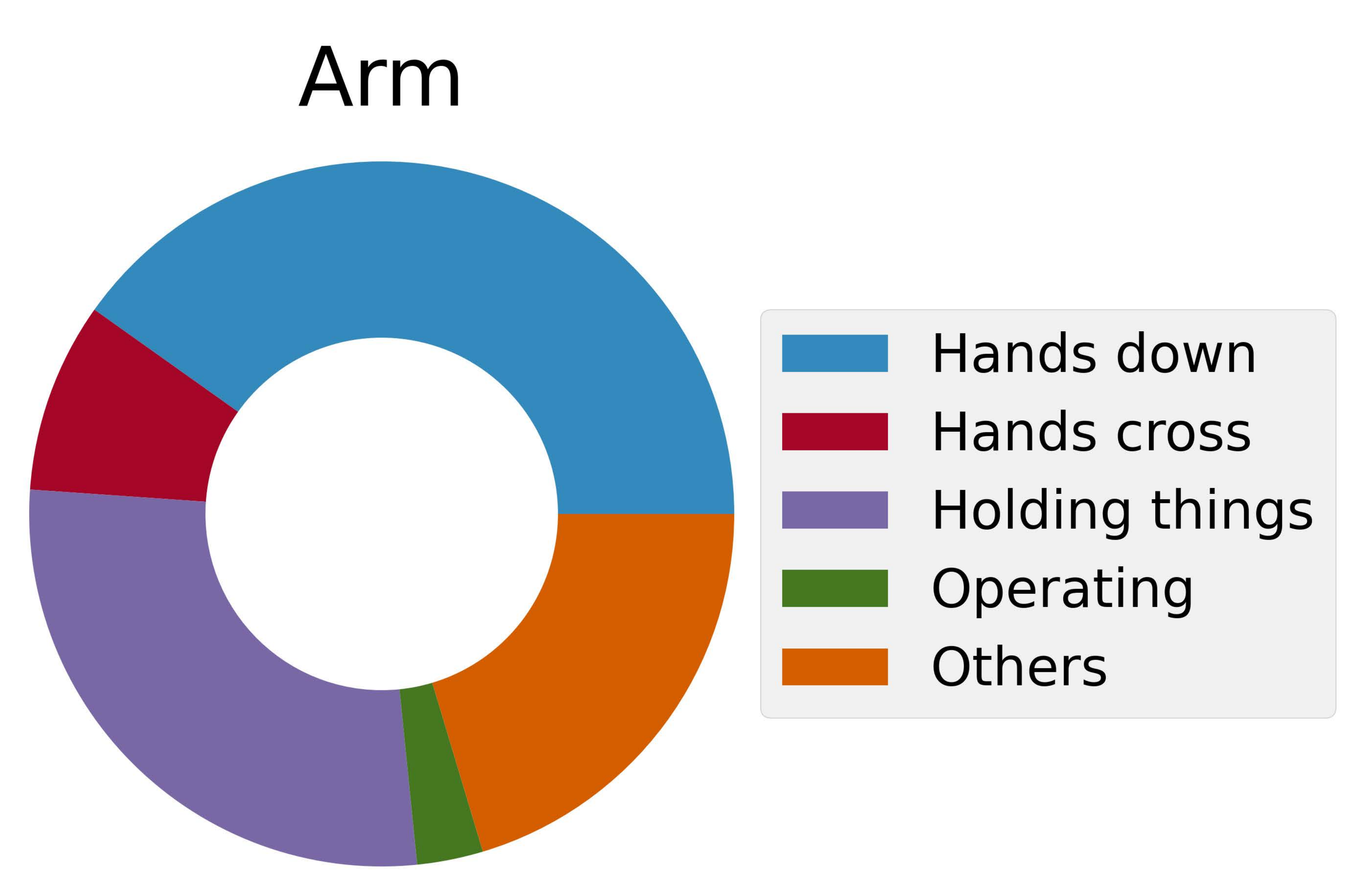}
  \includegraphics[width=0.24\textwidth]{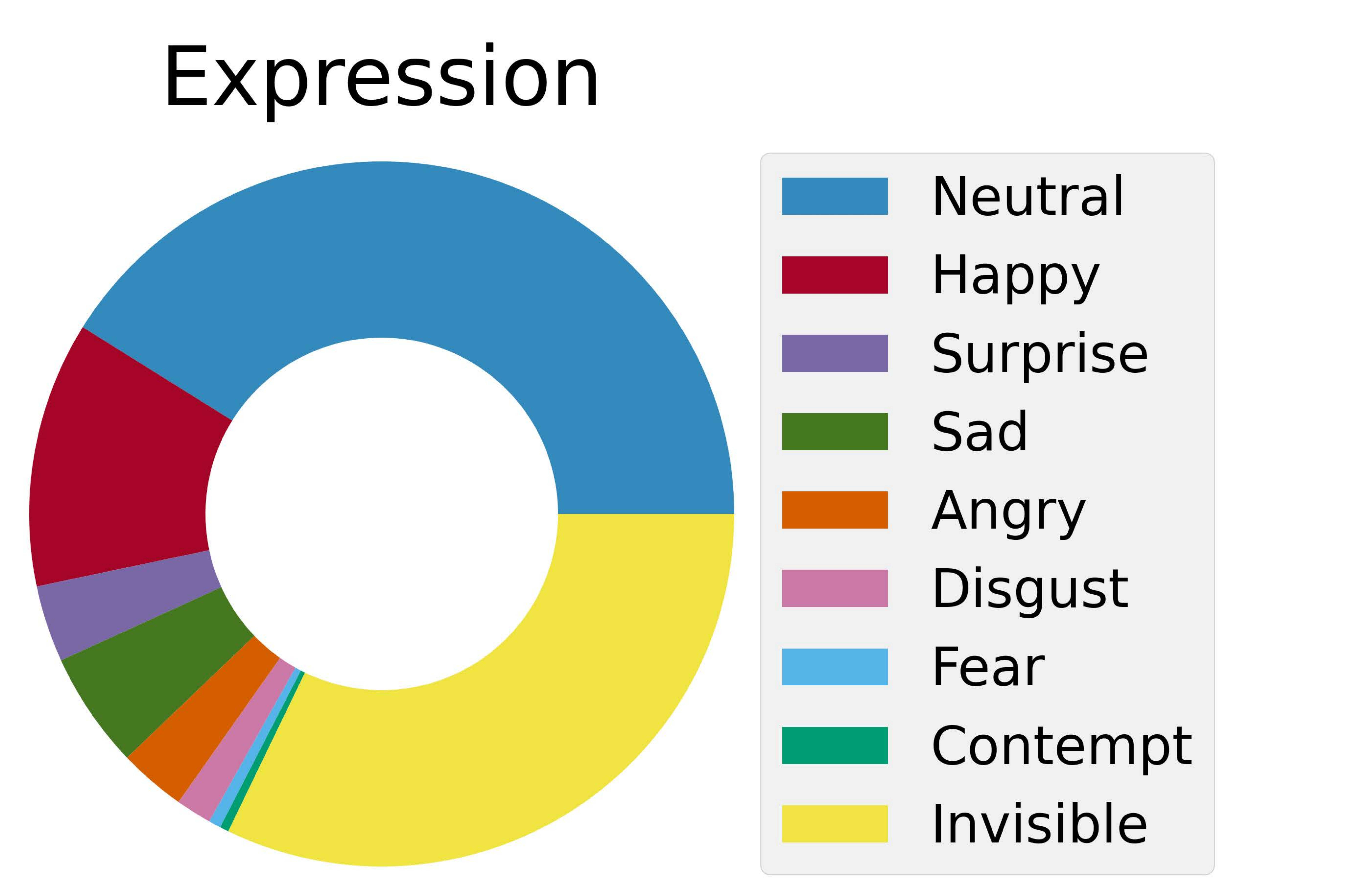}
  \caption{The distributions of samples among each label of attributes in Portrait250K.}
  \label{fig:info}
\end{figure*}

To facilitate the evaluation of models, we split the dataset. In particular, to meet the requirement of the re-id task, the IDs contained in the training set and the test set should not intersect, and the IDs contained in the query set should be a subset of the IDs in the gallery set. The sizes of each subset are shown in Table \ref{tab:split}.

\begin{table}
  \centering
  \begin{tabular}{p{1.3cm}<{\raggedright}|p{1.7cm}<{\raggedleft}p{1.7cm}<{\raggedleft}}
    \toprule
    Subset & \# of Images & \# of IDs \\
    \midrule
    Train & 151,633 & 640 \\
    Query & 12,361 & 404 \\
    Gallery & 86,006 & 426 \\
    \bottomrule
  \end{tabular}
  \caption{Sizes of each subset of Portrait250K.}
  \label{tab:split}
\end{table}

\subsection{Statistics and Features}

\noindent\textbf{Semi-Supervised ID.}
There are 86,516 and 57,724 images in the training set and gallery set respectively that have no ID labels, accounting for more than 60\% of the total. Most of them are not unrecognizable but do not belong to an important role, so the annotators didn't give them ID labels.

\noindent\textbf{Multi-Labeled Expression.}
Due to the complexity of facial expressions, annotators gave multiple expression labels to a small number of images, which makes the expression classification task here a multi-label problem.

\noindent\textbf{Long-Tailed and Unbalanced Distribution.}
We notice that the number of images owned by each character presents a significant long-tailed distribution and the distributions of samples among each label of other attributes are also seriously unbalanced (see Figure \ref{fig:info} for details of distributions). Previous work like \cite{cui2019class} uses the imbalance factor which is the number of training samples in the largest class divided by the smallest to indicate the severity of a long-tailed distribution, while in our dataset because the long-tailed distribution is naturally occurring rather than artificially constructed, the imbalance factor will ignore the other categories except for extreme cases, thus has no reference value. The Gini coefficient is used in economics to judge the fairness of income, but it cannot describe the distributions in more detail. We design the following indicator named ${\rm LTS}_k$ (Long Tail Score of $k$ proportion) to measure the severity of a long-tailed distribution: 
\begin{equation}
  {\rm LTS}_k=\frac{\mathop{\min}\limits_{\boldsymbol{x}^T\boldsymbol{y}\geq k\cdot \lVert \boldsymbol{y}\rVert_1} \lVert \boldsymbol{x}\rVert_1}{k\cdot N}
  \label{eq:lts}
\end{equation}
where $\boldsymbol{x}$ is a boolean vector of length $N$ and $\boldsymbol{y}$ records the number of samples corresponding to each label. $N$ is the number of labels, 1-norm represents the sum of elements and $k \in (0,1)$. ${\rm LTS}_k$ indicates the degree of enrichment of the largest classes that can account for $k$ proportion of all samples. The closer the value is to $0$, the more serious is the enrichment and on the contrary, the closer the value is to $1$, the more uniform is the distribution. We calculate ${\rm LTS}_{0.2}$ of the benchmark dataset Market-1501 \cite{zheng2015scalable} in the field of person re-id together with our Portrait250K, results are shown in Table \ref{tab:long-tail-compare} and obviously, Portrait250K has more serious long-tailed distribution. This is also true for other re-id datasets because they are all collected in a similar way. 

\begin{table}
  \centering
  \begin{tabular}{p{1.3cm}<{\raggedright}|p{2cm}<{\centering}p{2cm}<{\centering}}
    \toprule
    Subset & Portrait250K & Market-1501 \\
    \midrule
    Train & 0.086 & 0.400 \\
    Query & 0.074 & 0.766 \\
    Gallery & 0.070 & 0.446 \\
    \bottomrule
  \end{tabular}
  \caption{Comparison between Portrait250K and Market-1501 \cite{zheng2015scalable} on ${\rm LTS}_{0.2}$, lower values indicate more severe long-tailed distribution.}
  \label{tab:long-tail-compare}
\end{table}

From the image's point of view, the dataset also has the following two features which reflect situations that exist in the real world and increase the dataset's diversity and difficulty. 

\begin{figure}[t]
  \centering
  \includegraphics[width=0.2\linewidth, height=0.4\linewidth]{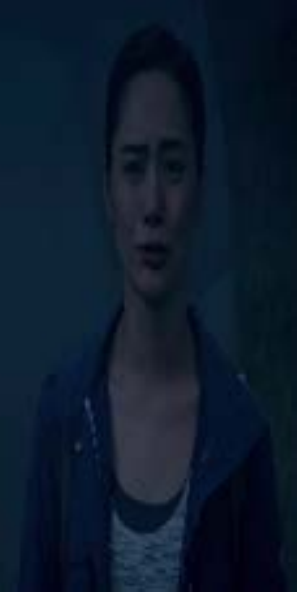}
  \includegraphics[width=0.2\linewidth, height=0.4\linewidth]{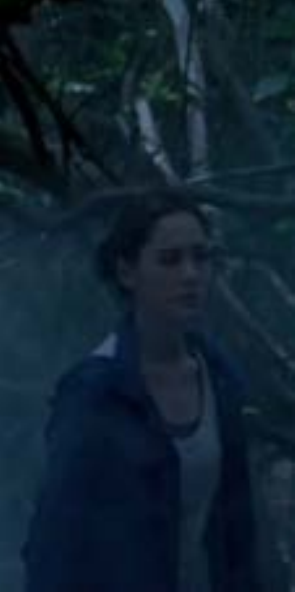}
  \includegraphics[width=0.2\linewidth, height=0.4\linewidth]{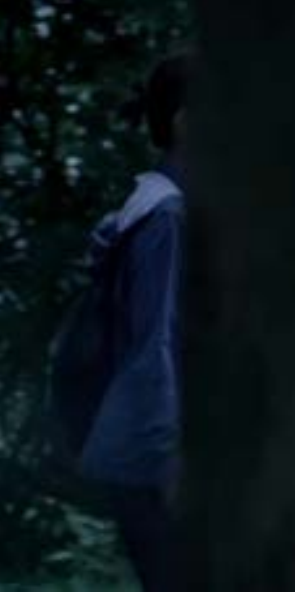}
  \includegraphics[width=0.2\linewidth, height=0.4\linewidth]{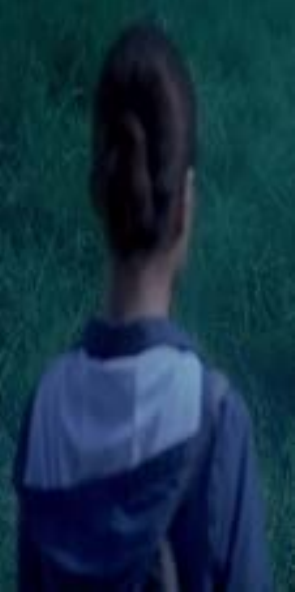}
  \caption{From \textit{left} to \textit{right}: the same person on the front, half side, side (with her right part occluded), and back.}
  \label{fig:occ}
\end{figure}

\noindent\textbf{Broader and More Diverse Occlusion, Truncation and Illumination.}
As shown in Figure \ref{fig:occ}, there are situations where the same person appears in different states or is occluded/truncated in different ways. There are also various weather conditions and time of shooting, resulting in varying illumination.

\begin{figure}[t]
  \centering
  \includegraphics[width=0.2\linewidth, height=0.4\linewidth]{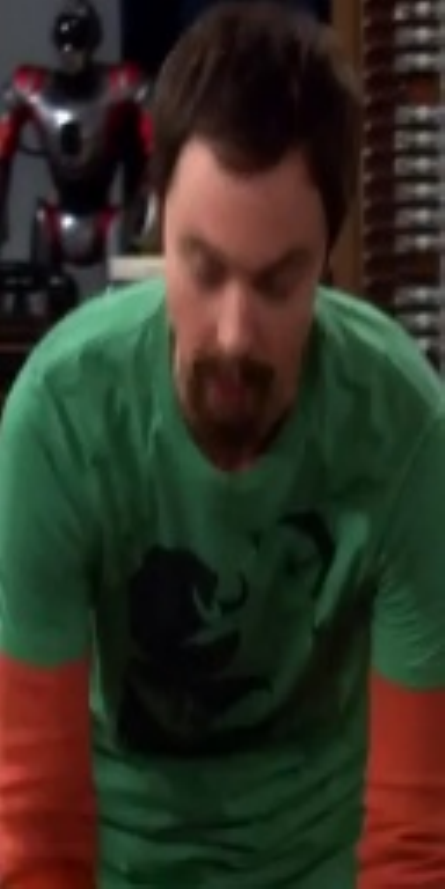}
  \includegraphics[width=0.2\linewidth, height=0.4\linewidth]{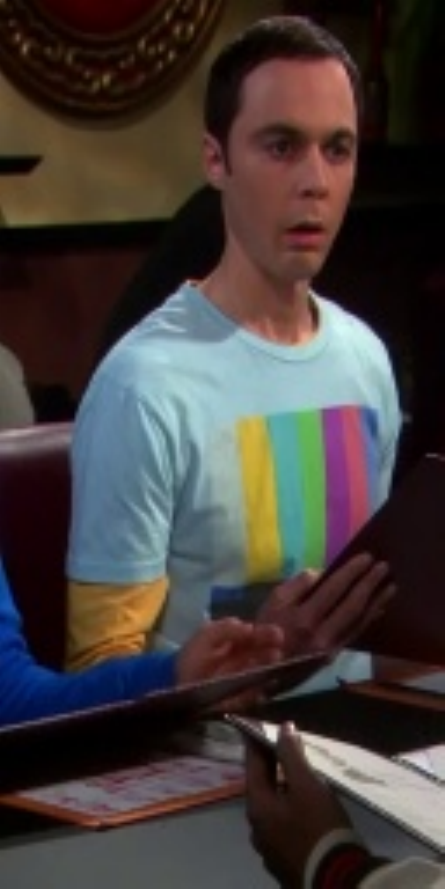}
  \includegraphics[width=0.2\linewidth, height=0.4\linewidth]{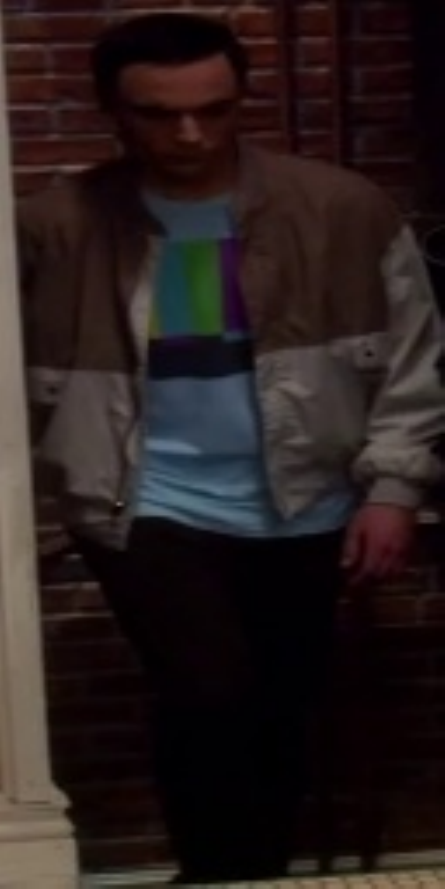}
  \includegraphics[width=0.2\linewidth, height=0.4\linewidth]{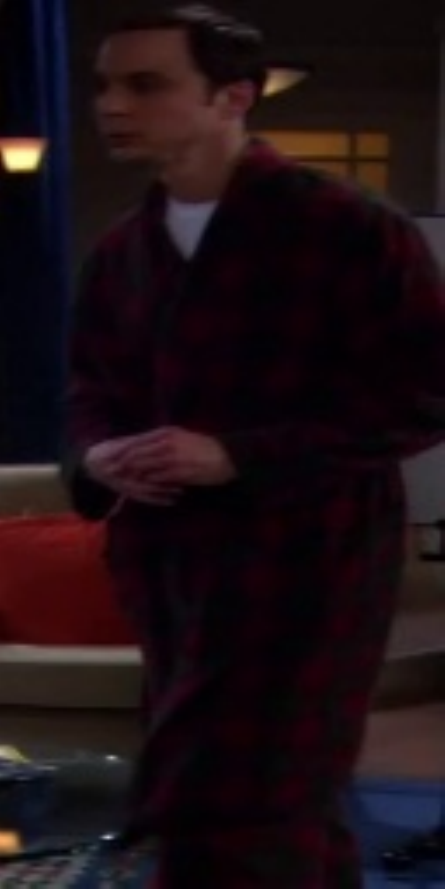}
  \caption{The same person has changes in clothing and makeup, and appears in different background environments.}
  \label{fig:change-look}
\end{figure}

\noindent\textbf{Clothing, Makeup and Background Changes.}
As shown in Figure \ref{fig:change-look}, there are situations where the same person has changes in clothing and makeup, and appears in a variety of different background environments. 

\subsection{Evaluation Metric}
\label{sec:metric}

For the simplicity of performance evaluation and comparison, and to highlight the original intention of portrait interpretation, we propose a unified metric that integrates metrics for all eight tasks to evaluate the overall performance of models, which we name Portrait Interpretation Quality (PIQ). Next, we first introduce metrics for each sub-task of portrait interpretation and then describe the computation of PIQ based on the metrics for each sub-task.

\noindent\textbf{Re-id Metrics.}
Commonly used evaluation metrics for person re-id are Cumulative Matching Characteristics (CMC) and mean Average Precision (mAP), and they are both computed by averaging the performance of all samples in the query set. This is rather reasonable when samples in the query set are nearly evenly distributed among each ID, which is also the actual situation of commonly used person re-id datasets like Market-1501 \cite{zheng2015scalable} (see Table \ref{tab:long-tail-compare}). 

However, when it comes to more general cases including Portrait250K, where the distribution of samples on different IDs in a large-scale query set is unbalanced, the above metrics will increase the weight of the head IDs. If query images are uniformly sampled among IDs, the size of the query set will be limited by the size of the tail IDs, resulting in an insufficient test set. In response to this problem, we propose Macro CMC and Macro mAP for Portrait250K. The difference is that they will first calculate the average performance of the samples of each ID in the query set, and then average among all IDs.

The task of re-id aims at performing cross-domain target retrieval. However, due to the fixed shot technique commonly used in film shooting, even if we select only one frame among several consecutive frames, there will be many similar frames. Similar frames do not have much impact on tasks other than re-id. Compared with removing them, retaining these similar frames is somewhat equivalent to data enhancement. But for the re-id task, if there exist similar frames of any query image in the gallery set, the query will become a simple sample since the model only needs to map similar images into close points of the embedding space. This will affect the effectiveness of the evaluation. For other re-id datasets like \cite{zheng2015scalable}, they naturally have no such problem since their test sets are composed of images taken by different cameras. 

To solve this problem, we need to mark similar images. Here we use the perceptual hash algorithm \cite{zauner2010implementation}, which can generate a hash value for each image, and the similarity of two images can be measured by calculating the Hamming distance of their corresponding hash values. We mark the images with similarities that exceed a threshold as a group and do not consider the gallery images in the same group as the query during testing. 

\noindent\textbf{Classification Metrics.}
For classification tasks, due to the unbalanced distribution of samples on different labels, we use the F1-score which is the harmonic average of precision and recall for evaluation. Similar to Macro CMC and Macro mAP, we use Macro F1-score.

\noindent\textbf{PIQ Metric.}
Gender, age, physique, and height classification are tasks regard to appearance, body and arm action classification are tasks regard to posture. The unified metric PIQ considers every aspects of the portrait equally, that is, the same weight is assigned to each aspect. However, it should be noted that although the re-id task belongs to appearance perception, it’s relatively independent and important, so we separate it. Naturally, we get the following formula for PIQ:
\begin{equation}
  {\rm PIQ=\frac{ReID.+App.+Pos.+Emo.}{4}}
  \label{eq:piq}
\end{equation}
where $\rm ReID.$ represents metric for the re-id task which is $\rm (Macro\ Rank1+Macro\ mAP)/2$ (we use Macro CMC at $\rm Rank1$ for evaluation), $\rm App.$ represents metric for appearance recognition which is $\rm (F1_{gender}+F1_{age}+F1_{physique}+F1_{height})/4$, $\rm Pos.$ represents metric for posture recognition which is $\rm (F1_{body}+F1_{arm})/2$ and $\rm Emo.$ represents metric for emotion recognition which is $\rm F1_{expression}$.
\section{Baseline Method}
\label{sec:model}

\begin{figure}[t]
  \centering
  \includegraphics[width=1.0\linewidth]{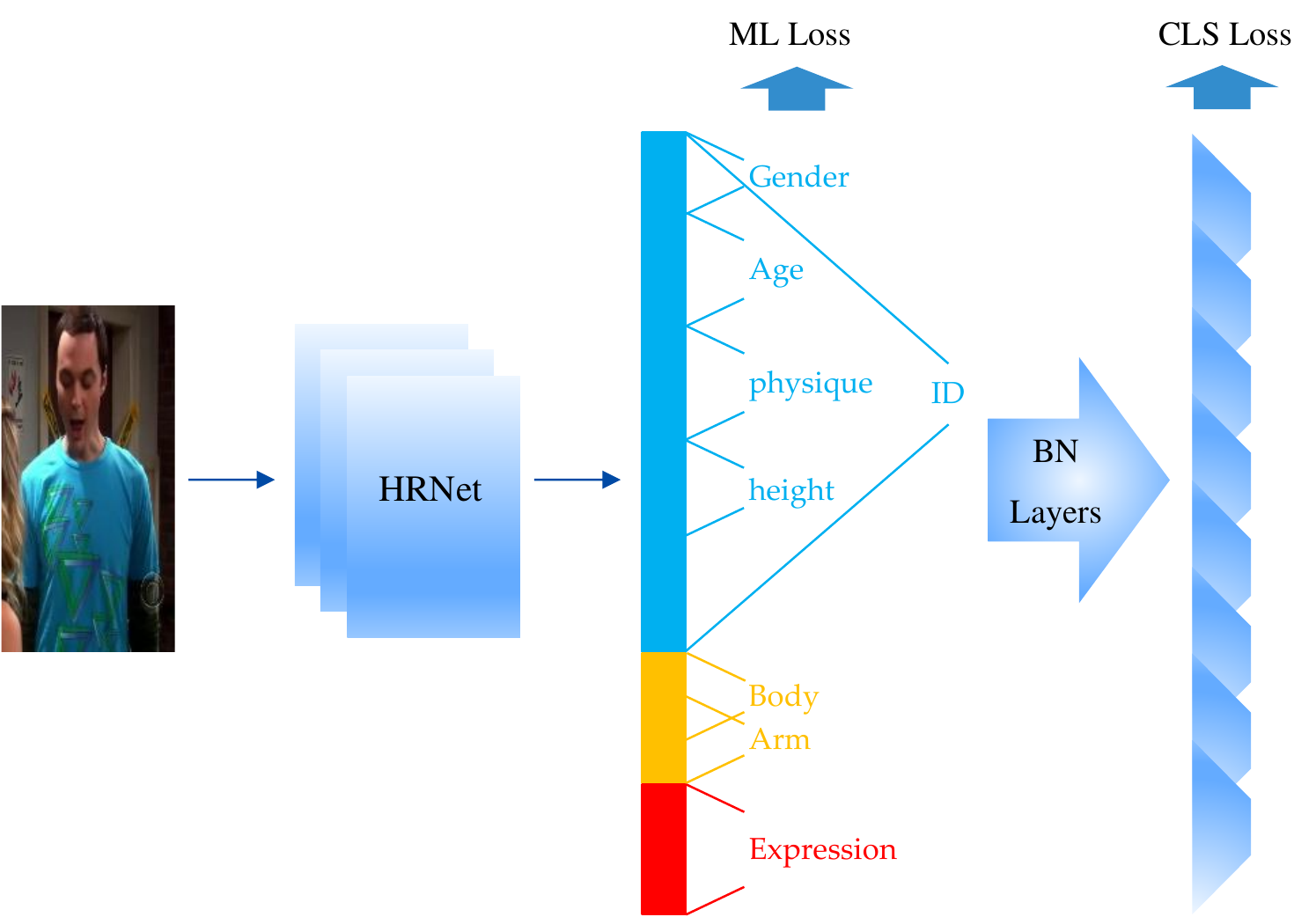}
  \caption{Illustration of our baseline method. The feature vector output by the HRNet backbone is divided following the relationship between aspects and tasks. The feature vectors for each task here are used to calculate the metric learning loss, and then the feature vectors for classifiers are obtained through BNNecks which are independent for each task.}
  \label{fig:net}
\end{figure}

We propose a baseline method for portrait interpretation, which is shown in Figure \ref{fig:net}. The proposed framework uses HRNet-W32 \cite{sun2019deep,wang2020deep} as a unified feature extractor, and the obtained feature vector is divided according to the three aspects of appearance, posture, and emotion. Different feature vectors are provided for the corresponding classifiers of each task. It also uses BNNecks \cite{luo2019bag,luo2019strong} with metric learning loss to improve the discrimination between categories.

\noindent\textbf{Feature Space Split.}
We realize that representation for the three aspects of appearance, posture, and emotion of portraits are naturally independent, so we divide the feature space into three sub-spaces to store information from the three aspects respectively. This claim is easy to establish because people with the same appearance (same person) can obviously make different postures or have different emotions, and vice versa. In practice, we directly divide the feature vector output by the backbone into three parts. For each sub-task, gender, age, physique, and height classification belong to appearance perception, but they are not related to each other, so they will correspond to disjoint parts of appearance's feature vector. The re-id task only focuses on identity-related information, which is exactly the information related to appearance. So the feature vector used by re-id is the feature vector assigned to appearance, which includes dimensions for related sub-tasks and some dimensions that are not assigned to any sub-tasks. By introducing information about various attributes, the re-id task can be guided by more supervision and thus improve performance. For posture, body and arm action classification are relatively related tasks, so we let the two tasks share some dimensions while each has its own dimensions. Emotion perception has only one sub-task so there is no need for further division.

For a classification task, more labels generally require a more complex feature space to make them distinguishable, so we let the number of dimensions of the feature vector corresponding to each task proportional to the number of related labels. 

\noindent\textbf{Loss Function.}
Metric learning can help the model learn better representations and has many applications in the fields of fine-grained image classification, face recognition, and re-id, etc. By utilizing metric learning losses, samples of the same category are pulled closer in the feature space, otherwise, they are pushed away. Most studies in metric learning use Euclidean distance as the distance metric, while classification losses like cross-entropy loss are optimized for the cosine distance. But in practice, we find that normalizing the feature vectors for metric learning can effectively improve performance, and the Euclidean distance is actually equivalent to the cosine distance in this case:
\begin{equation}
\begin{aligned}
  \lVert \boldsymbol{a}-\boldsymbol{b} \rVert_2&=\sqrt{\left(\boldsymbol{a}-\boldsymbol{b}\right)^T\left(\boldsymbol{a}-\boldsymbol{b}\right)} \\
  &=\sqrt{\boldsymbol{a}^T\boldsymbol{a}+\boldsymbol{b}^T\boldsymbol{b}-\boldsymbol{a}^T\boldsymbol{b}-\boldsymbol{b}^T\boldsymbol{a}} \\
  &=\sqrt{2-2\times \cos(\boldsymbol{a},\boldsymbol{b})}
  \label{eq:euc_cos}
\end{aligned}
\end{equation}

When we use normalized features for metric learning, in fact, the difference between this and the classification layer is that the latter optimizes for the cosine distance of each sample to the proxy (corresponding weight vector) of each category, while metric learning optimizes for the cosine distance between samples. Experiments indicate that metric learning can improve the robustness of representations and has the potential to generally improve classification performance. Among various metric learning losses, triplet loss \cite{schroff2015facenet,hermans2017defense} is one of the most frequently used, but it may suffer from slow convergence due to the large proportion of trivial triplets or noise samples when hard-mining is considered. We propose a metric learning loss named Batch Rank (BR) Loss that directly uses each sample in the batch as the anchor and optimizes for the distance from other samples to it: 
\begin{equation}
  \mathcal{L}_{\rm BR}=\frac{1}{N}\sum_{i=1}^{N}\sum_{\substack{j=1\\j\neq i\\y_i=y_j}}^{N} {{-\log \frac{\exp \left(\cos (f_i,f_j)\right)}{\sum_{\substack{c=1\\c\neq i}}^{N}\exp \left(\cos (f_i,f_c)\right)}}}
  \label{eq:rank}
\end{equation}
where $y$ denotes labels and $N$ is the batch size. BR loss gets rid of the common dependence on hyperparameters and can make full use of the given labels while reducing the impact of noise samples. A similar idea has also been mentioned by \cite{cao2007learning}.

Losses are calculated for the eight tasks separately, including the classification loss and the metric learning loss, and we summarize these two losses in a fixed proportion for each task:
\begin{equation}
  \mathcal{L}_i=\mathcal{L}_{i,cls}+\lambda \mathcal{L}_{i,ml}
  \label{eq:l_task}
\end{equation}
where $\mathcal{L}_i$ denotes loss for each task, $i=1\cdots8$.

We adopt an automatic learning scheme for weights of losses of each task as proposed in \cite{kendall2018multi} by using task-independent uncertainty. The overall learning objective is written as:
\begin{equation}
  \mathcal{L}=\sum_{i=1}^{8} {\frac{1}{e^{s_i}}\mathcal{L}_{i}+s_i}
  \label{eq:total_loss}
\end{equation}
where $s_i$ indicates the task-dependent uncertainty for each individual loss and are learnable parameters.

\noindent\textbf{Sampling Strategy.}
As far as the re-id task is concerned, to enable triplet loss with samples of moderate difficulty, \cite{hermans2017defense} proposes to randomly sample $P$ classes (identities) and $K$ images per class to constitute a training batch. But when the dataset contains a large number of unidentified samples (but labeled with other attributes), this sampling strategy will just ignore them. A simple solution is to mix the batch composed of the remaining unidentified samples from the $P$-$K$ sampling with the batch obtained from the $P$-$K$ sampling and randomly shuffle the order of them. But this may also lead to unstable supervision for the re-id task. Through experiments, we confirm that completely random sampling can obtain the best performance. 
\section{Experiments}
\label{sec:experiments}

In this section, we present experimental results with our Portrait250K dataset. We are especially interested in two questions: 1) Whether modeling the proposed portrait interpretation task in a multi-task paradigm is effective while efficient? 2) Whether the methods we propose, especially with regard to the idea of feature space split are effective?

\subsection{Implementation Details}

Our backbone networks are all pre-trained on ImageNet \cite{deng2009imagenet}. Data augmentation methods include random cropping, random horizontal flipping, and random erasing. The batch size is set to $128$. We use the Adam optimizer \cite{kingma2014adam}, and the base learning rate is set as $3.5e-4$, while we spend the first $10$ epochs to linearly warm it up from $3.5e-5$ following \cite{luo2019bag,luo2019strong} and drop it to $3.5e-5$, $3.5e-6$ and $3.5e-7$ at the 30th, 60th and 90th epochs, respectively. The training process is terminated at the 100th epoch.

\begin{table*}
  \centering
  \begin{adjustbox}{max width=\textwidth}
  \begin{tabular}{llll|cc|cccc|cc|c|c}
    \toprule
    \multirow{2}{*}{Model} & ML & MTL & Sampling & \multicolumn{2}{c|}{\textbf{Re-ID}} & \multicolumn{4}{c|}{\textbf{Appearance}} & \multicolumn{2}{c|}{\textbf{Posture}} & \textbf{Emotion} & \multirow{2}{*}{\textbf{PIQ}} \\
    & Loss & Loss & Strategy & mAP & Rank-1 & Gender & Age & Physique & Height & Body & Arm & Expression &  \\
    \midrule
    Single-Task & None & Ave. & random & {\color{red}{0.424}} & {\color{red}{0.603}} & {\color{red}{0.885}} & 0.717 & 0.287 & 0.615 & 0.527 & {\color{red}{0.634}} & {\color{blue}{0.202}} & {\color{blue}{0.480}}  \\
    Sim-MTL & None & Ave. & random & 0.314 & 0.509 & 0.847 & {\color{blue}{0.722}} & {\color{blue}{0.472}} & {\color{red}{0.708}} & {\color{red}{0.624}} & 0.596 & 0.183 & 0.473  \\
    FSS & None & Ave. & random & {\color{blue}{0.367}} & {\color{blue}{0.541}} & 0.831 & 0.718 & 0.471 & 0.692 & {\color{blue}{0.618}} & 0.598 & 0.176 & 0.479  \\
    \midrule
    FSS & Triplet & Ave. & random & 0.342 & 0.525 & 0.852 & 0.688 & 0.430 & 0.680 & 0.581 & 0.559 & 0.164 & 0.457  \\
    FSS & BR & Ave. & random & 0.359 & 0.538 & 0.837 & {\color{red}{0.740}} & {\color{red}{0.478}} & 0.699 & 0.617 & {\color{blue}{0.600}} & 0.174 & {\color{blue}{0.480}}  \\
    \midrule
    FSS & BR & Uncer. & random & 0.351 & 0.536 & 0.823 & 0.685 & 0.444 & 0.650 & 0.561 & 0.530 & {\color{red}{0.340}} & {\color{red}{0.495}}  \\
    \midrule
    Sim-MTL & None & Ave. & P-K & 0.310 & 0.514 & {\color{blue}{0.861}} & 0.474 & 0.279 & 0.525 & 0.448 & 0.412 & 0.181 & 0.389  \\
    Sim-MTL & None & Ave. & shuffle & 0.248 & 0.448 & 0.832 & 0.678 & 0.424 & 0.635 & 0.568 & 0.545 & 0.152 & 0.425  \\
    \midrule
    Sim-MTL-R50 & None & Ave. & random & 0.265 & 0.447 & 0.811 & 0.715 & {\color{blue}{0.472}} & {\color{blue}{0.700}} & 0.600 & 0.584 & 0.189 & 0.453  \\
    \bottomrule
  \end{tabular}
  \end{adjustbox}
  \caption{Experimental results under various conditions of the baseline method. Sim-MTL=Simple Multi-Task Learning baseline introduced in Section \ref{sec:mtl}, FSS=the proposed Feature Space Split method, Sim-MTL-R50=Simple Multi-Task Learning baseline with ResNet-50 \cite{he2016deep} as the backbone, Ave.=Average loss of multi-task, Uncer.=Uncertainty weighted loss proposed in \cite{kendall2018multi}, shuffle=shuffle the batch composed of the remaining unidentified samples from the $P$-$K$ sampling with the batch obtained from the $P$-$K$ sampling. Metrics for each task are Macro mAP, Macro Rank-1, and Macro F1 as discussed in Section \ref{sec:metric}.}
  \label{tab:exp}
\end{table*}

\subsection{Effectiveness of MTL and Feature Space Split}
\label{sec:mtl}

We design a simple multi-task learning baseline for comparison, which is a weakened version of the proposed method: the classification layer of each sub-task uses the same feature vector as input. We also conduct experiments when the same network solves a single task at a time, the only difference is that only the loss of a certain task is considered. For a fair comparison, we do not use the metric learning loss, and in the case of multi-task learning, the multi-task losses are averaged as the total loss. The relevant results are shown in the first part of Table \ref{tab:exp}.

As far as PIQ is concerned, the single-task models perform best, but the simple multi-task baseline can also achieve comparable performance. The feature space split method further reduces the gap greatly, which validates our idea. For each sub-task, we observe that: 1) Different models have the biggest performance gap on the re-id task. 2) Multi-task models perform significantly better on the three tasks of physique, height, and body action recognition, which means they can benefit greatly from supervision brought by labels for other tasks. 

\subsection{Influence of Loss Function}

\noindent\textbf{Influence of Metric Learning Loss.}
We test the triplet loss and the proposed batch rank loss on the model that adopts the feature space split method. The ratio of the metric learning loss to the classification loss is set to $0.5:1$. The relevant results are shown in the second part of Table \ref{tab:exp}.

We find that: 1) The proposed batch rank loss achieves better performance, and triplet loss may be affected by noise samples. 2) The performance of the re-id task decreases with the introduction of metric learning losses, which is probably because  the $P$-$K$ sampling is not used. In that case, there are not many cases where samples with the same ID appears in each batch (128 is used as the batch size, while there are 640 people in the training set), which makes other tasks capable of calculating stable metric learning losses more dominant in training. 

\noindent\textbf{Influence of Multi-Task Loss.}
When we adopt the uncertainty weighted loss proposed in \cite{kendall2018multi}, the model performance is further improved, shown in the third part of Table \ref{tab:exp}. This is because the model is greatly improved on the most difficult facial expression classification task. 

\subsection{Supplementary Experiments}

\begin{table}
  \centering
  \small
  \begin{tabular}{l|ccc}
    \toprule
    ML Loss & mAP & Rank-1 & Rank-5 \\
    \midrule
    None & 0.871 & 0.951 & {\color{blue}{0.985}} \\
    Triplet & {\color{red}{0.886}} & {\color{blue}{0.955}} & {\color{red}{0.987}} \\
    BR & {\color{blue}{0.885}} & {\color{red}{0.956}} & 0.983 \\
    \bottomrule
  \end{tabular}
  \caption{Experiments on the person re-id task about the metric learning loss. Conducted on Market-1501 \cite{zheng2015scalable}.}
  \label{tab:reid-ml}
\end{table}

\noindent\textbf{Sampling Strategy.}
To verify the superiority of random sampling, we also conduct experiments with $P$-$K$ sampling and the method of mixing $P$-$K$ sampling and random sampling. Results are shown in the fourth part of Table \ref{tab:exp}, showing that random sampling performs the best. 

\noindent\textbf{Backbone Network.}
We also test the baseline when replacing its backbone with the commonly used ResNet-50 \cite{he2016deep} (close in scale), the experimental result is shown in the last part of Table \ref{tab:exp}. Thanks to reliable high-resolution representations, HRNet-W32 \cite{sun2019deep,wang2020deep} can get better performance. 

\noindent\textbf{ML Loss on re-id task.}
To show the versatility of the batch rank loss, we also conduct experiments on the person re-id task. We use the commonly used Market-1501 \cite{zheng2015scalable} dataset and adopt HRNet-W32 as the baseline, with triplet loss and our batch rank loss, respectively. Note that the feature vectors for triplet loss calculating are normalized for better performance. The results are comparable, shown in Table \ref{tab:reid-ml}.
\section{Conclusion and Discussion}
\label{sec:conclusion}

In this paper, we propose the task of Portrait Interpretation and construct a large-scale benchmark dataset for it. We also propose a baseline method. Portrait Interpretation systematically divides the perception of humans into Appearance, Posture, and Emotion. Through several sub-tasks in each aspect, models are required to give portraits a comprehensive description of static attributes and dynamic states. The dataset we construct contains 250,000 images with eight kinds of annotations which are sufficient enough. Through experiments, we prove that the multi-task learning-based scheme can achieve performance beyond the single-task baseline on this task, which further proves the feasibility and superiority of the proposed task. 

\noindent\textbf{Future of Portrait Interpretation.}
The sub-tasks included in portrait interpretation can be changed flexibly. For example, the more difficult task of pose estimation \cite{toshev2014deeppose} can be used to represent the posture aspect. Facing the real world, researches on long-tailed or unbalanced recognition should also be introduced. Self- \& semi-supervised learning can make use of massive unlabeled data, and can also be incorporated into the training process of related models. Furthermore, models trained for portrait interpretation are expected to serve as general pre-trained models, and their ability to solve other tasks after fine-tuning is also worth exploring. 

\noindent\textbf{Potential negative societal impact.}
Similar to tasks such as re-id, the training of a powerful portrait interpretation model may require more annotated data, and the collection and use of these data pose the risk of privacy leakage. Existing solutions include federated learning \cite{mcmahan2017communication} and so on. 


{\small
\bibliographystyle{ieee_fullname}
\bibliography{egbib}
}

\appendix

\section{Details of Portrait250K}

\subsection{More statistics}

\begin{figure*}[t]
  \centering
  \includegraphics[width=1.0\textwidth]{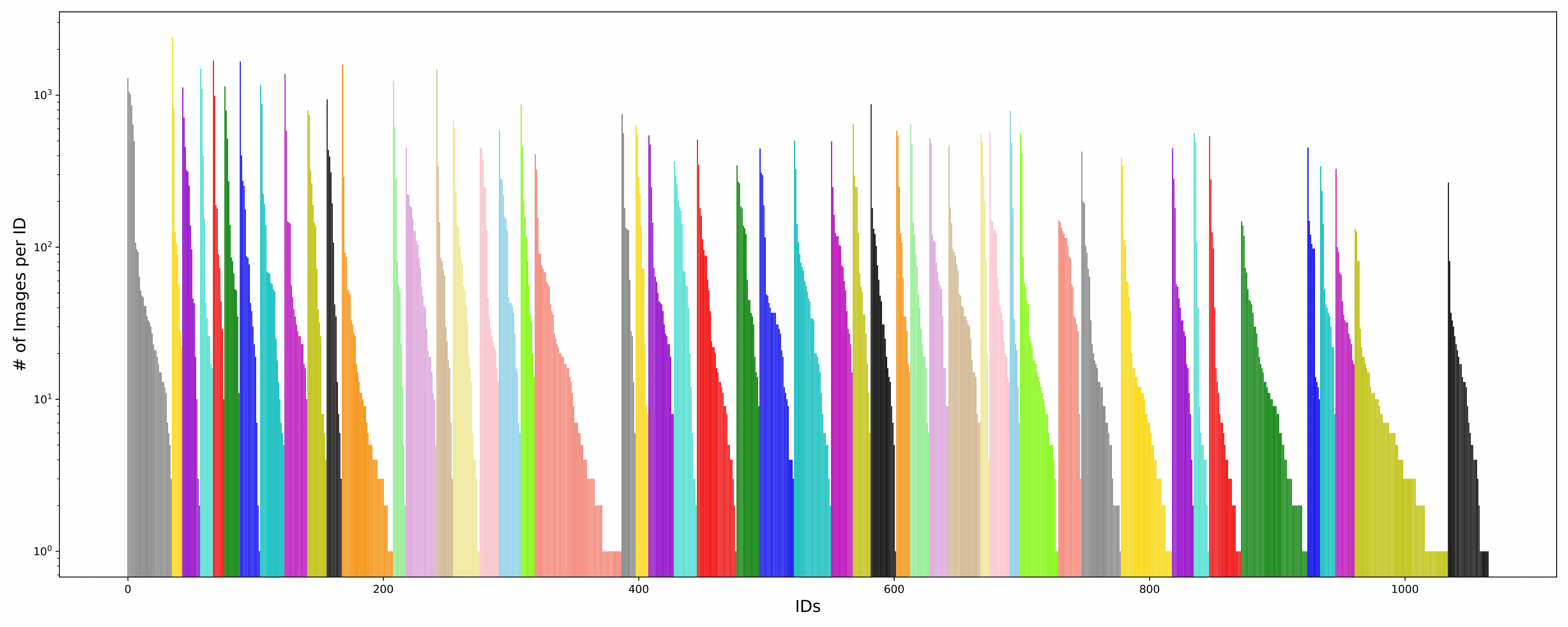}
  \caption{The distribution of samples among each ID in each movie/TV series in Portrait250K.}
  \label{fig:id}
\end{figure*}

As introduced in the paper, our dataset has 1,066 IDs from 51 movies and TV series. Table \ref{tab:data_s} shows our data source, which covers extensive diversity. Figure \ref{fig:id} shows the distribution of samples among different IDs, different colors represent different movies or TV series. From left to right, the total numbers of images from each movie/TV series are in descending order. 

\begin{table*}
  \centering
  \footnotesize
  \begin{threeparttable}
  \begin{tabular}{p{0.23\textwidth}<{\raggedright}p{0.1\textwidth}<{\centering}p{0.08\textwidth}<{\centering}|p{0.23\textwidth}<{\raggedright}p{0.1\textwidth}<{\centering}p{0.08\textwidth}<{\centering}}
    \toprule
    Movies/TV Series & Release Year & IMDb & Movies/TV Series & Release Year & IMDb \\ \midrule
    \multicolumn{6}{c}{\textbf{United States}} \\ \midrule
    Song One & 2014 & tt2182972 & Braveheart & 1995 & tt0112573 \\
    Knucklehead & 2010 & tt1524131 & Line of Duty & 2013 & tt1532538 \\
    Broken Promise & 2006 & tt0810786 & Hush Money & 2017 & tt4828304 \\
    The Good Doctor & 2011 & tt1582271 & 37 & 2016 & tt4882174 \\
    Management & 2008 & tt1082853 & Unconditional & 2011 & tt1758610 \\
    The Matador & 2005 & tt0365485 & Friends \tnote{1} & 1994 & tt0108778 \\
    The Big Bang Theory \tnote{2} & 2009 & tt1487706 & Bad Boys for Life & 2020 & tt1502397 \\
    Flipped & 2010 & tt0817177 \\ \midrule
    \multicolumn{6}{c}{\textbf{Canada}} \\ \midrule
    Bailout: The Age of Greed & 2013 & tt2368553 \\ \midrule
    \multicolumn{6}{c}{\textbf{United Kingdom}} \\ \midrule
    Brexit: The Uncivil War & 2019 & tt8425058 & Sorry We Missed You & 2019 & tt8359816 \\
    The Gentlemen & 2019 & tt8367814 \\ \midrule
    \multicolumn{6}{c}{\textbf{Japan}} \\ \midrule
    Being Good & 2015 & tt3818826 & Sway & 2006 & tt0809535 \\
    Goodbye, Grandpa! & 2017 & tt7009956 & South Bound & 2007 & tt1073545 \\
    Journey to the Shore & 2015 & tt4173478 \\ \midrule
    \multicolumn{6}{c}{\textbf{South Korea}} \\ \midrule
    Inside Men & 2015 & tt3779028 & My New Sassy Girl & 2016 & tt4080594 \\
    Office & 2015 & tt4682562 & A Violent Prosecutor & 2016 & tt5442308 \\
    Season of Good Rain & 2009 & tt1477859 \\ \midrule
    \multicolumn{6}{c}{\textbf{Thailand}} \\ \midrule
    A Little Thing Called Love & 2010 & tt1859438 & Teng Nong Khon Maha-Hia & 2007 & tt1058014 \\
    Seasons Change & 2006 & tt0880477 & Super Salaryman & 2012 & tt2754800 \\
    Heart Attack & 2015 & tt4964598 \\ \midrule
    \multicolumn{6}{c}{\textbf{China}} \\ \midrule
    Drug War & 2012 & tt2165735 & Crazy New Year’s Eve & 2015 & tt4481934 \\
    My Best Friend’s Wedding & 2016 & tt5275314 & The Grandmaster & 2013 & tt1462900 \\
    Gone with the Time & 2015 & tt6450032 & Almost a Comedy & 2019 & tt11273332 \\
    The Winners & 2020 & tt11563836 & Dying to Survive & 2018 & tt7362036 \\
    Yip Man 4 & 2019 & tt2076298 & Missing & 2019 & tt12043620 \\ \midrule
    \multicolumn{6}{c}{\textbf{India}} \\ \midrule
    3 Idiots & 2009 & tt1187043 & Andhadhun & 2018 & tt8108198 \\ \midrule
    \multicolumn{6}{c}{\textbf{Germany}} \\ \midrule
    Miss Sixty & 2014 & tt3478536 & Victoria & 2015 & tt4226388 \\ \midrule
    \multicolumn{6}{c}{\textbf{France}} \\ \midrule
    Sky & 2015 & tt4106306 \\ \midrule
    \multicolumn{6}{c}{\textbf{Netherlands}} \\ \midrule
    The Surprise & 2015 & tt3409440 \\ \midrule
    \multicolumn{6}{c}{\textbf{Spain}} \\ \midrule
    The Invisible Guest & 2016 & tt4857264 \\
    \bottomrule
  \end{tabular}
  \begin{tablenotes}
    \footnotesize
    \item[1] S01E01-E04
    \item[2] S03E01, S03E13
  \end{tablenotes}    
  \end{threeparttable} 
  \caption{Movies and TV series used by Portrait250K.}
  \label{tab:data_s}
\end{table*}

\subsection{Annotation Interface}

\begin{figure*}[t]
  \centering
  \includegraphics[width=1.0\textwidth]{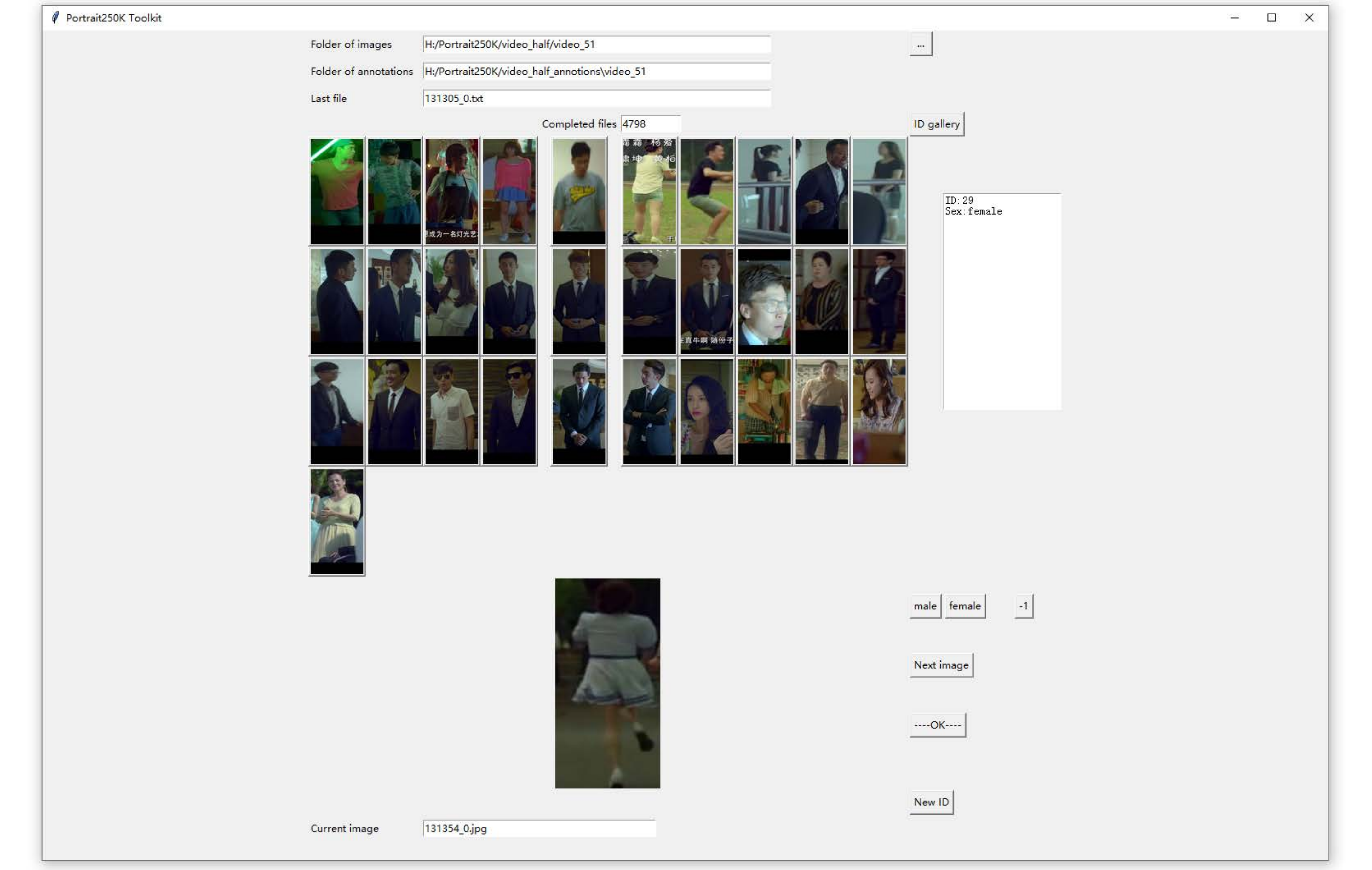}
  \caption{Annotation interface for ID and gender.}
  \label{fig:tool}
\end{figure*}

To improve the efficiency and accuracy of annotation, we need to provide our annotators with suitable tools, existing practices include reCAPTCHA \cite{von2008recaptcha}. Thus we design a set of toolkits for annotation. Among the eight kinds of labels, ID is the most difficult one for annotators. Figure \ref{fig:tool} shows the interface of the system we designed when annotating identity and gender. The software will display an example image for each of the characters that have been recognized in the movie. The annotator only needs to click on the example image that matches the current image (displayed in a larger size at the bottom of the interface) to select its identity. And together the gender label is selected (because the same person has a constant gender). When the current image does not match any of the example images, the annotator needs to click to create a new identity, and then select a gender label. 

\section{License}
Please notice that the Portrait250K dataset will be made available for academic research purpose only. The copyright of all the images belongs to the original owners. 


\end{document}